\documentclass[conference]{IEEEtran}
\IEEEoverridecommandlockouts
\usepackage{cite}
\usepackage{amsmath,amssymb,amsfonts}
\usepackage{algorithmic}
\usepackage{graphicx}
\usepackage{textcomp}
\usepackage{xcolor}
\usepackage{comment}
\usepackage{algorithm}
\usepackage{subfigure}
\usepackage{soul}
\usepackage{fancyhdr}
\def\BibTeX{{\rm B\kern-.05em{\sc i\kern-.025em b}\kern-.08em
    T\kern-.1667em\lower.7ex\hbox{E}\kern-.125emX}}

\usepackage[top=0.72in, bottom=0.98in, left=0.625in, right=0.625in]{geometry}

\begin{document}

\pagestyle{fancy}
\fancyhead[C]{\textbf{Best Paper Award}: IEEE/ACM International Conference on Big Data Computing, Applications and Technologies (BDCAT) 2022}
\title{Detection of Uncertainty in Exceedance of Threshold (DUET): An Adversarial Patch Localizer\\

}



\author{\IEEEauthorblockN{Terence Jie Chua$^{\mathrm{1}}$, Wenhan Yu$^{\mathrm{1}}$, Chang Liu$^{\mathrm{1}}$, and Jun Zhao$^{\mathrm{2}}$} \\
\IEEEauthorblockA{ $^{\mathrm{1}}$Graduate College, Nanyang Technological University (NTU) \\
       $^{\mathrm{2}}$School of Computer Science and Engineering, Nanyang Technological University (NTU)  
       }}

\maketitle
\thispagestyle{fancy}
\begin{abstract}
Development of defenses against physical world attacks such as adversarial patches is gaining traction within the research community. We contribute to the field of adversarial patch detection by introducing an uncertainty-based adversarial patch localizer which localizes adversarial patch on an image, permitting post-processing patch-avoidance or patch-reconstruction. We quantify our prediction uncertainties with the development of \textit{\textbf{D}etection of \textbf{U}ncertainties in the \textbf{E}xceedance of \textbf{T}hreshold} (DUET) algorithm. This algorithm provides a framework to ascertain confidence in the adversarial patch localization, which is essential for safety-sensitive applications such as self-driving cars and medical imaging. We conducted experiments on localizing adversarial patches and found our proposed DUET model outperforms baseline models. We then conduct further analyses on our choice of model priors and the adoption of Bayesian Neural Networks in different layers within our model architecture. We found that isometric gaussian priors in Bayesian Neural Networks are suitable for patch localization tasks and the presence of Bayesian layers in the earlier neural network blocks facilitates top-end localization performance, while Bayesian layers added in the later neural network blocks contribute to better model generalization. We then propose two different well-performing models to tackle different use cases.
\end{abstract}


\section{Introduction}
In recent years, deep learning have been widely adopted across both academia and industry. However, rapid adoption of deep learning techniques attracts adversaries. These adversaries typically attempt to attack or fool machine learning models by introducing adversarial inputs. In the field of computer vision, attacks are most commonly acted on the image or video input~\cite{xiong2021multi}. The attacks could be acted on the pixels, or on their corresponding label(s). Many industries rely heavily on deep learning technologies to automate or to enhance the performance of tasks. Taking autonomous vehicles as an example, these cars utilize computer vision technologies which empower self-driving capabilities. An adversarial attack on the machine's perceived image could skew its comprehension and hence, prediction of the scene~\cite{xiong2021multi}. An attack as such could result in disastrous accidents. Taking deep-learning enhanced medical imaging as another example, attacks on medical images or the deep learning model could alter the perception of the machine~\cite{lee2019physical}. A malign object could be classified and perceived as benign. The consequence of such an attack could result in incorrect understanding of a patient's medical condition.



Many of these attacks~\cite{madry2017towards} leveraged on the existence of highly vulnerable parts of an image or video perceived by a machine, and make alterations to pixels in these regions. However, these picture-wide attacks are hard to implement as attackers usually do not have access to machine's perceptions. Due to the impracticality of these picture-wide attacks on images in the real world, there has been an increase in focus on real-world physical attacks. Real-world physical attacks refer to an adversary implementing attackers on real-world objects~\cite{lee2019physical}. Using the same autonomous vehicle example to emphasize our point: a physical attacker could be an adversarial sticker placed on road-side signboards with the aim of fooling deep learning models mounted on autonomous vehicles to perceive the sign incorrectly.

The devastating consequences of real-world physical attacks and their successes have also inspired developments in adversarial defences. There are 3 main types of adversarial defences that have been proposed and developed across the recent years. They are (i) adversarial training~\cite{madry2017towards}, (ii) robustification of models~\cite{wong2018provable}, and (iii) adversarial detection~\cite{deng2021libre}. Adversarial training is the training of machine learning models with adversarial examples. On the other hand, adversarial robustification works aim to improve the inherent robustness of models to adversarial examples without involving any training on adversarial examples~\cite{wong2018provable}. Although adversarial training and adversarial robustification are widely studied defence approaches, they often suffer the issue of having poorer clean data input performance. Adversarial detection does not suffer the issue of poorer clean data input performance as they often detect and sieve out adversarial examples or objects prior to being fed into deep learning models.

\begin{figure}[htb]
\vspace*{-0.5cm}
\centering
\includegraphics[width=0.30\textwidth]{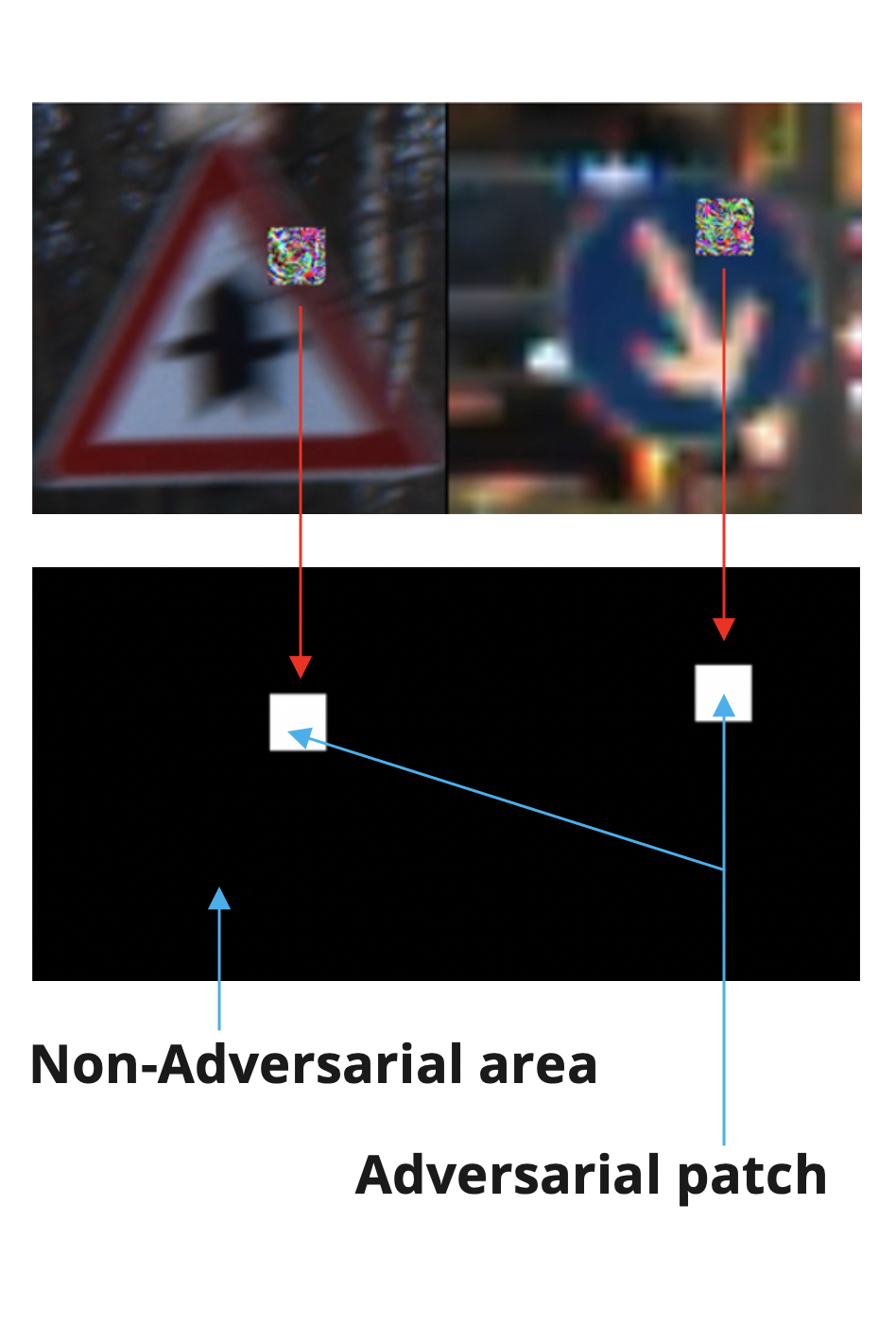}
\vspace{-1cm}
\caption{Localization of adversarial patch}
\label{Introduction_adversarial}
\vspace*{-0.5cm}
\end{figure}

Based on our current knowledge of adversarial defences, we developed a physical-world uncertainty-based adversarial patch localizer. This patch localizer aims to not only identify the presence of adversarial patches but also localize the patches through segmentation methods (shown in Figure \ref{Introduction_adversarial}). Localizing adversarial patches is important in adversarial patch detection so that post-detection patch-reconstruction or patch-avoidance can be performed. In addition to developing an adversarial patch localizer, we quantify our prediction uncertainties with the development of \textit{\textbf{D}etection of \textbf{U}ncertainties in the \textbf{E}xceedance of \textbf{T}hreshold} (DUET) algorithm. Our adversarial patch localizer utilizes Bayesian neural networks (BNN) ~\cite{blundell2015weight} to provide us with uncertainty estimates of adversarial pixels and the DUET algorithm provides us with a framework to ascertain confidence in our predictions. In several safety-sensitive fields including medical imaging and autonomous vehicles, it is essential to ascertain with a specified amount of confidence that each pixel that the machine deems as non-adversarial, is indeed non-adversarial. On the flip side, any pixels that do not meet their required confidence will then be flagged as possible adversarial. Adversarial detectors have been postulated to perform poorly on out-of-distribution samples (samples the detector has not seen before). The adoption of BNNs in our adversarial patch localizer not only allows us to quantify the uncertainty of our prediction estimates, but has also shown to be able to generalize well and respond better than traditional point-estimate neural networks to unseen adversarial patches. To the best of our knowledge, there is no existing literature that focuses on localizing adversarial patches through uncertainty quantification and segmentation. Our work is novel and contributes to existing literature by:

\begin{itemize}
    \item Introducing a novel uncertainty-based adversarial patch localizer under our proposed DUET framework.

    \item Showcasing the importance and superiority of Bayesian neural networks over traditional point-estimate neural networks in adversarial patch localizers.

    \item Providing in-depth analysis on the implementation of Bayesian neural layers in adversarial patch localizers.
\end{itemize}

\section{Related works}
\textbf{Adversarial Attacks. }
In recent years, many gradient-based white-box and alternative forms of black-box model attacks have been proposed in the field of adversarial attacks. Most of the white-box attacks rely on information about the model's gradients and compute the attacks based on the gradients of the loss function with respect to the inputs~\cite{madry2017towards}. Of the many methods proposed, the most iconic and notable adversarial attacks are the $l_{\infty}$-norm, $l_2$-norm Projected Gradient Descent (PGD)~\cite{madry2017towards} and Fast-Gradient Sign Method (FGSM)~\cite{goodfellow2014explaining}. These attack algorithms produce perturbations by the sign of the gradient, and their perturbation sizes are controlled by the step size.

\textbf{Adversarial Detection. }
Earlier works in adversarial detection have been focused on detecting the presence of adversarial perturbations in images~\cite{ma2018characterizing}. Many works built classifiers to identify adversarial images~\cite{feinman2017detecting}, while several other works~\cite{ma2018characterizing,yang2020ml} studied the inherent statistical properties of adversarial and natural examples and leveraged on these properties to make a prediction.

\textbf{Patch detection. }
As physical attack's practicality in real-world gains recognition, there is an increasing number of researches focused on developing better defenses against adversarial patch attacks. Authors of~\cite{brown2017adversarial} were pioneers in this sub-field and showcased an attempt at crafting adversarial patches to successfully fool classification models. Recent adversarial patch detection works~\cite{arvinte2020detecting,xu2020lance} have adopted heuristic approaches to their detection algorithms. Existing heuristics-based methods include leveraging the difference between wavelet-filtered images from the original images in their adversarial patch detection~\cite{arvinte2020detecting}. Another example of a heuristics-based algorithm used in adversarial patch detection is the use of Grad-CAM maps to distinguish between natural and adversarial examples, identifying the existence of adversarial patches~\cite{xu2020lance}. However, these heuristics-based methods still lack robustness against strong adversarial attacks.

\textbf{Bayesian adversarial detectors. }
 The abovementioned works have proposed deterministic model architectures for their adversarial detectors. However, due to the deterministic nature of the weights, it renders these detector models vulnerable to fooling, especially if adversaries know the model weights. This weakness experienced by deterministic models has inspired works to adopt a bayesian approach in their model architecture, bayesian neural network~\cite{feinman2017detecting,lu2018limitation}. However, much of these works focused on building classification-based BNN detector~\cite{lu2018limitation}.

\textbf{Adversarial attack localization. }
There is little literature focused on localizing adversarial perturbations in an image using segmentation techniques. Work by~\cite{besnier2021triggering} has utilized a segmentation model to detect the presence of out-of-distribution (OOD) samples semantic maps through identification of adversarial samples. However, this work uses a semi-replicate version of the original segmentation model, coupled with the original model to produce a confidence map. After that, this confidence map is used to flag the presence of OOD samples. The study's main goal is to detect the presence of OOD samples and hence their models do not localize adversarial patches.

\section{Attack Formulation}
For our study, we assigned square-shaped patches of size smaller than 2\% of the total image area randomly on our training dataset. We adopted the $l_{\infty}$-norm PGD attack on the square patch as an example attack to be applied on our training set. The mechanism of PGD can be described as such: PGD algorithm aims to find a perturbation value to be added to the natural example, which maximizes the loss function value, under the constraints in which the norm of the perturbation value falls within a pre-defined threshold (shown in equation \ref{pgd_function}).
\vspace*{-0.5cm}

\begin{align}
    \max_{\lVert\zeta\rVert_\infty \leq \gamma} \ell(f(x_0 + \zeta; w), y_0)
    \label{pgd_function}
\end{align}

\noindent where $x_0$ represents the natural example, $y_0$ represents the original label, $\zeta$ is the perturbation value to be added, $w$ is the model weights, $\ell$ is the loss function, and $\gamma$ represents the perturbation threshold value. Implementing it by iterative gradient descent, we have:
\vspace*{-0.5cm}

\begin{align}
    x_{t+1} = x_{t} + \alpha \cdot \text{sign}(\nabla_x \ell(f(x_t;w),y_0)) \label{iterative}
\end{align}

\noindent where $x_t$ refers to the current image state, and $\alpha$ is a scalar multiplier.
Intuitively, the value of $x_{t+1}$ at the next iteration is the sum of the direction of the gradient of the loss function at the current step $x_t$, which is multiplied by a weight, and the current step $x_t$. To increase variety in our adversarial attacks, and to test model generalizability, we have included Google Patch and Fast Gradient Sign Method (FGSM) attacks on our dataset to be used during model testing. FGSM can be expressed similarly to equation \ref{iterative}. The main difference between FGSM and PGD is that PGD involves a random initialization where the starting point is random within an $l_{infinity}$-ball. On the other hand, Google Patch adds a different touch by localizing gradient calculations and updates to the patch itself which makes the attack within the patch stronger ~\cite{brown2017adversarial}. Note that each of the attacks implemented in our work was applied iteratively.

\section{Detection of Uncertainty in Exceedance of Threshold (DUET) framework}
\subsection{Model Architecture}
\begin{figure*}[htb]
\centering
\includegraphics[width=0.7\textwidth]{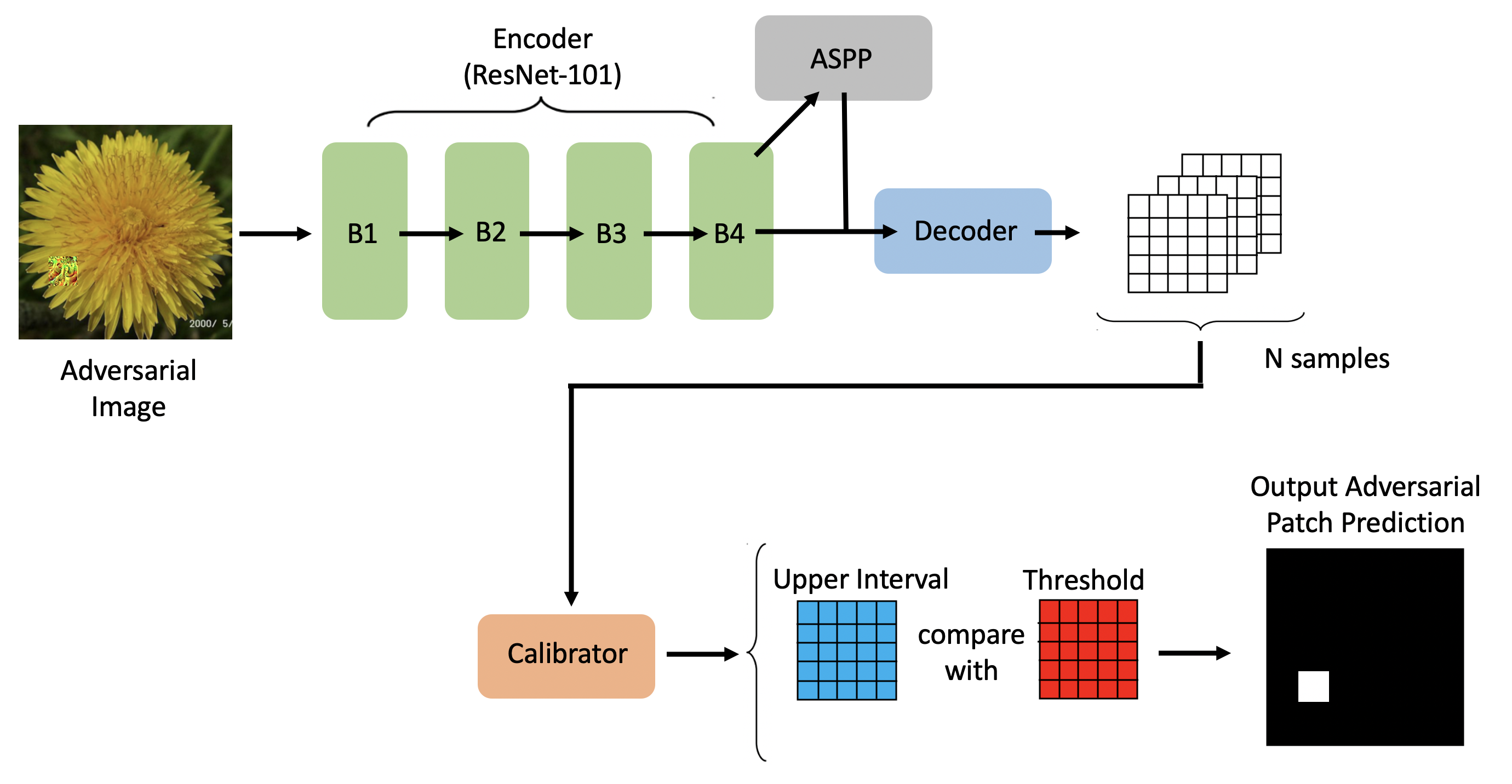}  
\vspace*{-0.5cm}
\caption{Adversarial Patch localizer under DUET framework. B1, B2, B3, B4 are the ResNet-101 blocks.}
\label{duet}
\vspace*{-0.35cm}
\end{figure*}

We adopted the $Deeplab V3+$ segmentation architecture~\cite{chen2018encoder} for our adversarial patch localizer. A segmentation model takes in an input image and outputs a feature map consisting of the class label for each pixel. The output of our adversarial patch localizer is a feature map, with $1s$ being predicted adversarial pixels and $0s$ being predicted non-adversarial pixels. We have adopted a pre-trained ResNet-101 backbone as the encoder. Attached to the end of the encoder is an \textit{Atrous Spatial Pyramid Pooling} (ASPP) module and the output feature maps from the backbone encoder and the ASPP is concatenated prior to being fed into the decoder. The decoder then produces the predicted output segmentation map. Note that the output carry values in the domain $[0,1]$ and can be loosely thought of as the probability that a pixel is adversarial. An output from the adversarial patch localizer will carry these values in a $M$x$M$ tensor, where $M$ is the height and width of the input image. Our DUET model architecture is shown in Figure \ref{duet}.

\subsection{Bayesian Neural Network}
The success of patch localizers is measured by its ability to identify and localize adversarial patches (of seen and unseen attack algorithms). However, it is infeasible to train adversarial patch localizers to identify attacks by all existing and potential adversarial algorithms. Bayesian neural network (BNN) is a natural choice for the localization of adversarial patches of varied attacks due to its ability to detect out-of-distribution (OOD) samples~\cite{liu2018adv}. Furthermore, as Bayesian neural networks incorporate stochasticity into the functional model, it offers greater model flexibility to generalize to varying attack signatures~\cite{liu2018adv}.

In training BNNs, the goal is to approximate the posterior over the weights $p(w|x,y)$ where $w$ are the weights of the neural network, $x$ are the input images, and $y$ are the labels. We are interested in finding the posterior over the weights $p(w|x,y)$ given a prior of the weights $p(w)$. We use the isometric gaussian distribution for the prior as it is commonly used for variational inference~\cite{liu2018adv}:
\vspace*{-0.5cm}

\begin{align}
    q_{\theta_i}(w_i) = \mathcal{N}(w_i;\mu_i;\sigma^2_i)
\end{align}

\noindent where $i$ represents the $i^{th}$ weight in the network. Based on the bayes rule, $p(w|x,y)$=$\frac{p(x,y|w)\cdot p(w)}{p(x,y)}$. However finding $p(x,y)$ is intractable as $p(x,y)$=$\int p(x,y|w) \cdot p(w) \mathrm{d}w$, resulting in infinitely large number of computations due to the continuous nature of $p(w)$ and the number of weights in the model. The integral above is a short-hand notation for the set of all $w$ in the model. A feasible method for bayesian inference, and the one we adopt for this work, is variational inference, where the true posterior $p(w|x,y)$ is approximated with a distribution $q_{\theta}(w)$. The idea is to learn the parameter $\theta$ of the approximating distribution which minimizes or provides the closest approximation to the exact posterior $P(w|x,y)$. The parameter $\theta$ can be estimated through minimizing the Kullback-Leibler (KL) divergence between the true posterior distribution $p(w|x,y)$ and the approximate distribution $q_{\theta}(w)$.
\vspace*{-0.75cm}

\begin{align}
    \theta^* = \text{arg min}_{\theta} KL\left(q_{\theta}(w)\lVert p(w|x,y)\right)
    \vspace*{-0.5cm}
\end{align}

\noindent where the $KL\left(q_{\theta}(w)\lVert p(w|x,y)\right)$ is:
\vspace*{-0.25cm}

\begin{align}
    KL\left(q_{\theta}(w)\lVert p(w|x,y)\right)=\int_{w}q_{\theta}(w)\text{log}\left( \frac{q_{\theta}(w)}{p(w|x,y)}\right)\mathrm{d}w
\end{align}

\noindent Note that this form still requires us to compute the $p(w|x,y)$. However, to overcome this issue, we can utilize the Evidence Lower Bound (ELBO)~\cite{hoffman2016elbo} formulation, by rewriting the above form:
\vspace*{-0.5cm}

\begin{align}
    \int_{w}q_{\theta}(w)&\log\left( \frac{q_{\theta}(w)}{p(w|x,y)}\right)\mathrm{d}w \nonumber\\&= -KL\left(q_{\theta}(w)\lVert p(w,(x,y))\right) + \text{log}(p(x,y))
\end{align}

\noindent This formulation is derived by using bayes rule, and the log$p(x,y)$, which is the evidence, does not depend on the weights. To simplify the notations further, we can rewrite the above equation as:
\begin{align}
    \int_{w}q_{\theta}(w)\text{log}\left( \frac{q_{\theta}(w)}{p(w|x,y)}\right)\mathrm{d}w = -\mathcal{L}(q) + \text{log}(p(x,y))
\end{align}

\noindent where $q$ represents the variational posterior and $p(x,y)$ is the marginal. $\mathcal{L}(q)$ is the lower bound of the evidence (ELBO). Due to the nature of the above equation, minimizing the KL-divergence between the variational posterior and the exact posterior requires the maximization of the ELBO:
\vspace*{-0.5cm}

\begin{align}
    q^*_{\theta}(w) = \text{arg max}_{q_{\theta} \in Q}(\mathcal{L}(q))
\end{align}

\noindent where $q_{\theta}^*$ is the optimal variational distribution, $q_{\theta}$ represent the set of variation distribution within set $Q$. We then utilized the stochastic variational inference algorithm to compute the ELBO for each mini-batch every iteration. We adopted normal distribution  for the approximating posterior and an isometric gaussian distribution for its prior, due to ease of computation, as it provides an elegant closed form KL-divergence solution~\cite{liu2018adv}. In implementing bayes-backprop, we obtain the stochastic parameter $\phi$ by adding noise following a normal distribution $\phi= \theta + \mathcal{N}(0,\sigma)$, where the noise is sampled at each iteration.

\textbf{Bayesian implementation. }We replace the encoder portion of the original \textit{Deeplab V3+} with Bayesian neural nets.

\textbf{Inference. } During test phase, monte carlo (MC) samples are drawn from the variational posterior $q_{\theta}(w)$. To produce an output probability distribution for epistemic uncertainty calibration, we take $N$ sets of $q_{\theta}(w)$ to compute the output probability feature maps $p(y|x,w)$ for each pixel in the image.

\subsection{Uncertainty Calibration}
A common way to produce confidence estimates for an output probability is to consider the confidence intervals of the output distribution. These intervals are typically calculated by taking specific quantiles of the output distribution, based on the required confidence. For our study, we set our confidence interval as 90\% ($5^{th}$ and $95^{th}$ quantile); i.e., we want to identify the window in which the 90\% adversarial patch localizer output probabilities interval contains the true probability value 90\% of the time. Nevertheless, since we are only concerned with our adversarial classification threshold value sitting below our tolerance window, we are essentially establishing that our threshold has to sit above our 95\% quantile, one-tail window for us to ascertain with 95\% confidence that a given predicted non-adversarial pixel is indeed non-adversarial. However, a direct extraction of the confidence interval from the outputs of a bayes model often results in an underestimation of the true confidence interval; i.e., the 90\% confidence interval do not contain the true output probability value 90\% of the time. The rationale for the underestimation is that variational inference methods often result in a misspecified model. Therefore, we calibrate the output confidence intervals via an extension of Platts Scaling~\cite{kuleshov2018accurate}. Note that since it is almost impossible to know the true probabilities for each pixel, we approximate the true probability values using only the output probabilities obtained by the correctly classified pixels of a non-bayes version of our proposed adversarial patch localizer.

\begin{figure}[htb]
\centering
\includegraphics[width=0.45\textwidth]{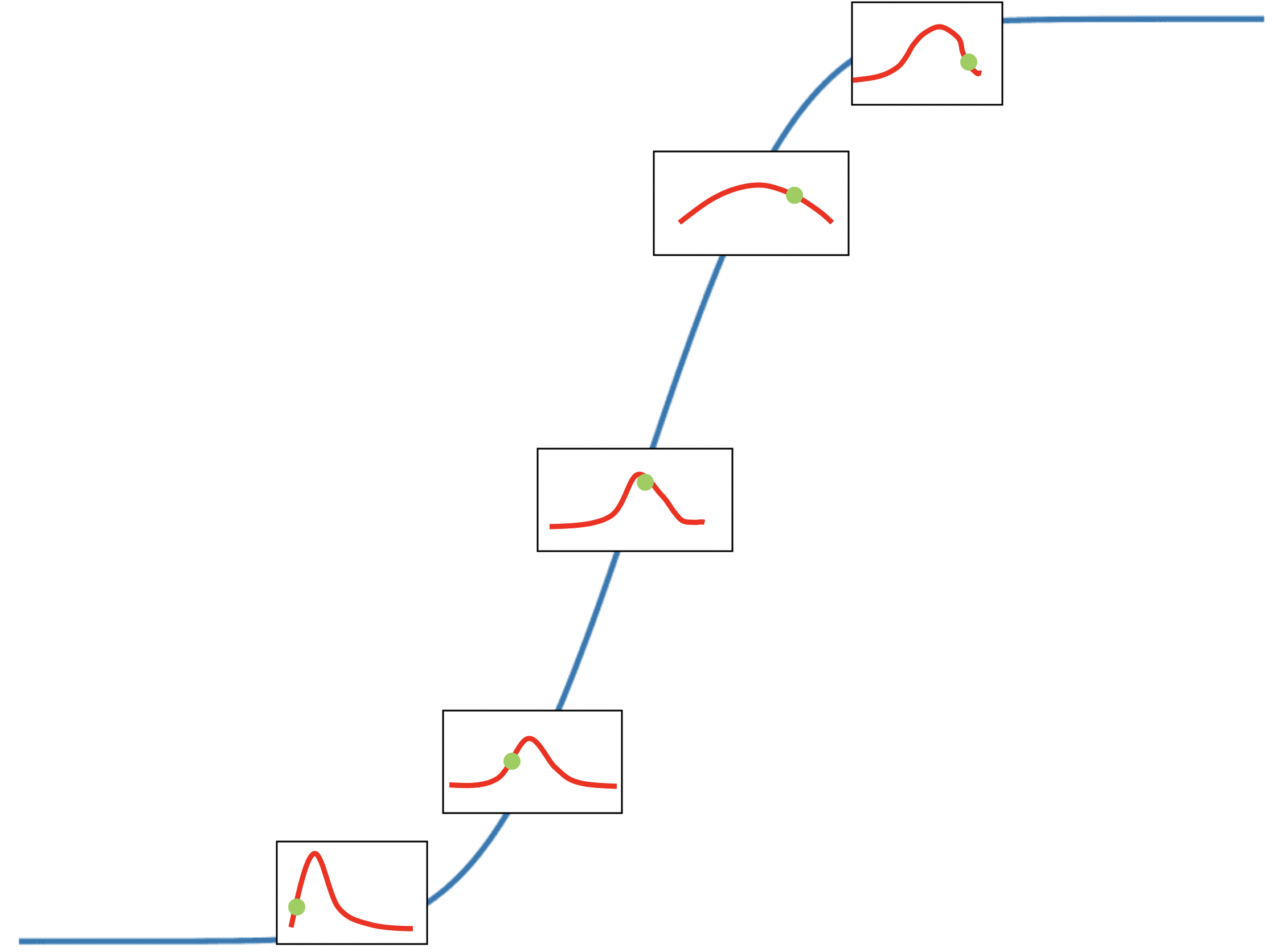}  
\caption{Illustration of true probability value for a pixel within in its predicted PDF (within small box), and true probability value's position in its own PDF in comparison to where other pixel’s true probability value lies in their PDF (box's position on CDF).}
\label{calibration}
\end{figure}

The intuition for this calibration is to identify where the true output probability value for each pixel estimate lies in the predicted output distribution of probabilities and compare its relative position against the relative positions of the true probability values of other pixels in their own predicted output probability distribution (shown in Fig. \ref{calibration}). This is to facilitate the mapping of the predicted output probability distribution at each point so that the calibrated distribution would be in line with the predicted distributions across other pixels. This mapping is performed to map the desired predicted interval to the predicted interval produced by the bayesian neural model. This is done by training a calibration model on a calibration train set and using the calibrated isotonic model to produce the calibrated confidence interval. To train a calibration model, we created a calibration dataset:
\begin{align}
    D=\left\{x,y\right\}
\end{align}
\noindent where $D$ represents the calibration dataset where value $x$ denotes where the true probability value for a pixel lie in its predicted probability density function (PDF) (shown as position of green circle on the PDF within the small boxes in Fig. \ref{calibration}). On the other hand, $y$ value for the corresponding  pixel denotes where its true probability value stand in its own PDF in comparison to where other pixel's true probability value lies in their PDF (shown as the position of the boxes on the cumulative density function (CDF) in Fig. \ref{calibration}). With both variables for each pixel, an isotonic regression model is fitted and is subsequently used for confidence interval calibration.

\begin{figure*}[htb]
\centering
\includegraphics[width=0.9\textwidth]{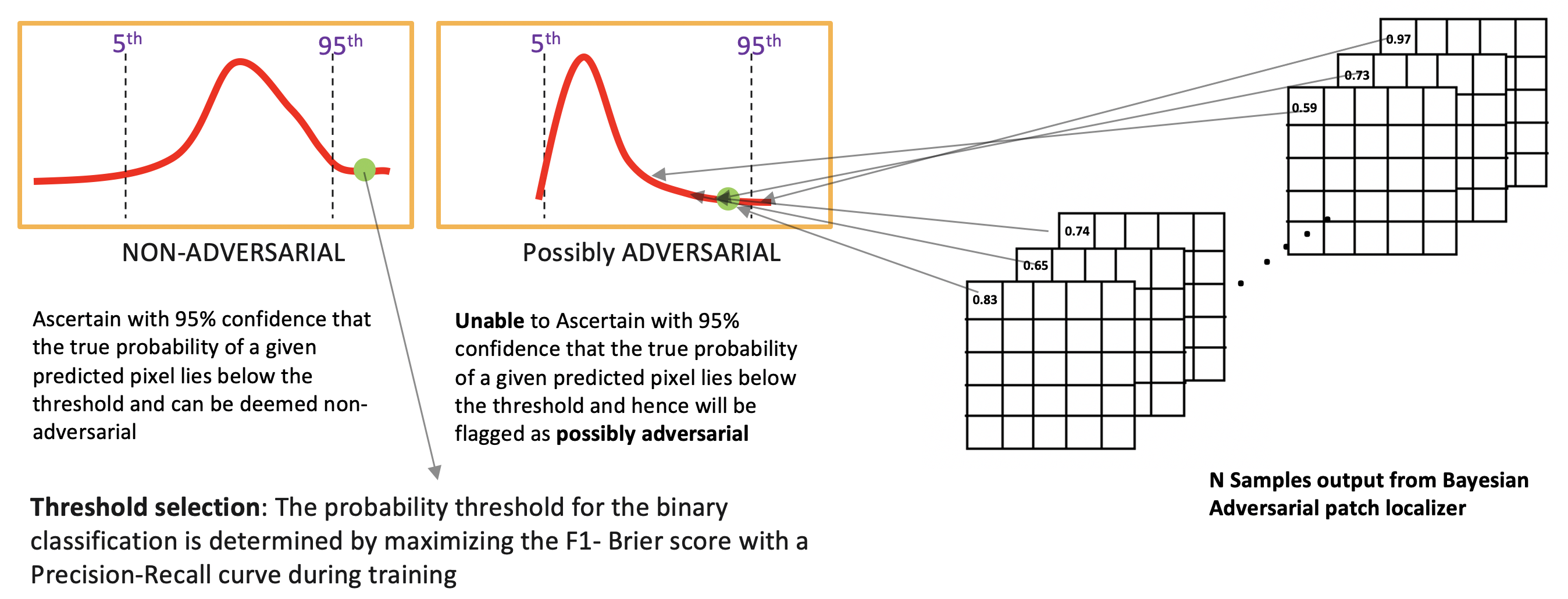}  
\caption{illustration of the classification of pixels under the DUET framework. Green dot represents the probability threshold selected, while the dotted lines refer to the $5^{th}$ and $95^{th}$ confidence interval values.}
\label{threshold_exceedance}
\vspace*{-0.35cm}
\end{figure*}

\subsection{Threshold Selection}
The probability threshold for the binary classification is determined by maximizing the F1-Brier score with a Precision-Recall curve \cite{davis2006relationship} during training.

\subsection{Exceedance of Threshold}
With the calibrated confidence interval for the Bayesian model output probability estimates, we propose the \textit{\textbf{D}etection of \textbf{U}ncertainties in \textbf{E}xceedance of \textbf{T}hreshold} (DUET) algorithm. The main goal of the algorithm is to detect if the adversarial binary classification threshold lies below the specified confidence interval. For our work, we set our required confidence interval as 90\% (or one-tail 95\%) as that is a generic mainstream required confidence for semi-safety-sensitive applications. If the upper window of the predicted probability confidence interval for a pixel in an image lies below the adversarial binary classification threshold, we can say that \textit{we are confident that the true predicted probability value does not lie above the adversarial binary classification threshold 95\% of the time}, and will be deemed as non-adversarial. In the case where the upper window of the 90\% confidence interval sits above the classification threshold, we cannot ascertain  that the true probability value sits below the adversarial binary threshold 95\% of the time and hence, the output pixel will be deemed as adversarial. Intuitively, the \textit{DUET} algorithm attempts to quantify uncertainty by ensuring that the allowable confidence interval (i.e., 90\%) of the probability estimates must fall below the classification threshold to be deemed as non-adversarial. Otherwise, the algorithm cannot ascertain with the allowable confidence interval that the probability estimate lies below the classification threshold, and may indeed be adversarial (illustrated in Fig. \ref{threshold_exceedance}). The \textit{DUET} algorithm is very applicable for safety-sensitive applications such as medical imaging and autonomous vehicles where there needs to be great confidence in classifying an input pixel as benign or non-adversarial but would not hesitate to raise or flag pixels as adversarial if the output probabilities do not fall within the acceptable confidence interval. Our DUET algorithm can be observed in Algorithm \ref{duetalgo}.

\begin{algorithm}[t] 
\caption{DUET algorithm} 
    \begin{algorithmic}[1] \label{duetalgo}
        \STATE {\textbf{Input: }image $x$, mask $y$, pre-trained BNN model $f(\cdot)$, calibrator $g(\cdot)$, confidence interval $c$}
        \STATE {\textbf{Output: }adversarial localization map $\hat{y}$}
        \STATE {\textbf{procedure } DUET ($f(\cdot)$, $g(\cdot)$, $x$, $y$, $c$, $T$)}
        \FOR{$i$ $\in$ range(sample count)}
            \STATE{$\hat{z}_i$ = $f(x,y)$}
            \STATE{Add $\hat{z}_i$ to $predList$}
        \ENDFOR
        \FOR{pixel $j,k$ in $m\cdot m$ output $\hat{z}$}

            \STATE{$upperWindow_{j,k}$ = $quantile \left(predList_{j,k}, c + \frac{100-c}{2}\right)$}
            \STATE{$calibratedUpperWindow_{j,k}$ = $g(upperWindow_{j,k})$}
            \IF{$calibratedUpperWindow_{j,k} \leq$ Threshold $T$}
                \STATE{pixel $j,k$ is \textbf{non-adversarial}}
            \ELSE
                \STATE{pixel $j,k$ is \textbf{adversarial}}
            \ENDIF
        \ENDFOR
        \RETURN{$\hat{y}$}

    \end{algorithmic}
    
\end{algorithm}

\section{Prior and Bayesian Layer Selection}
\subsection{Prior selection}
In Bayesian frameworks, the prior is an essential part in determining the performance of the model as the prior influences the posterior. In deep learning, a prior can be thought of as a regularizer~\cite{liu2018adv}, or a constraint on the model which prevents over-fitting of the model. In most works focused on Bayesian neural networks, the most common priors adopted is the isometric Gaussian distribution which acts similar to the $l_2$ regularization in traditional deep learning~\cite{liu2018adv}. In this work, we explore the use of other priors such as \textit{Cauchy} and \textit{Gaussian Mixture Models}  and explain the suitability or lack-of in our adversarial localizer. The Cauchy distribution is shown as: 
\vspace*{-0.25cm}

\begin{align}
    \text{Cauchy}(\kappa) = \frac{1}{\pi\kappa\left[1+\left(\frac{x}{\kappa}\right)^2\right]}
\end{align}

\noindent Cauchy distribution is chosen as a prior for us to test as it is much more efficient to compute. We attempt with a Gaussian Scale Mixture model as it provides much greater flexibility than conventional models. The Gaussian Scale Mixture model is as shown below:
\vspace*{-0.5cm}

\begin{align}
    \text{G}^M(\sigma_0,\sigma_1,\beta) = \beta\mathcal{N}(0,\sigma^2_0)+(1-\beta)\mathcal{N}(0,\sigma^2_1)
    \vspace*{-0.25cm} 
\end{align}

\noindent where $\mathcal{N}$ denotes the probability density function of the normal distribution, $\sigma$ is the standard deviation, and $\beta$ is the weighting assigned to each distribution.

\subsection{Bayesian Layer(s) Selection}
From previous works, we have observed works which use Bayesian neural nets in their deep learning architecture and incorporate them at vastly different layers in a model. In our work, we have incorporated the Bayesian layers in different parts of our segmentation architecture. We designed a fully-bayesian segmentation architecture (Encoder-Decoder-Bayes), encoder-only Bayes (Encoder-Bayes), specific encoder backbone block Bayes (Encoder-$x_0x_1x_2x_3$-Bayes), and decoder-only bayes (Decoder-Bayes). From here on, we refer to Encoder-$x_0x_1x_2x_3$-Bayes as a DUET model with $x_{i}$ block being bayes if $x_{i}$ is $1$.

\section{Experiment setting and results}
\subsection{Evaluation}

In this section, we conduct experiments to evaluate the prowess of our DUET-framework Adversarial Patch Localizer. As we aim to show how our proposed safety-sensitive bayesian adversarial patch localizer algorithm is suitable to be adapted on existing, underlying segmentation models, we compare our proposed bayesian-based Adversarial Patch Localizer with a non-stochastic segmentation model. Since we are interested in quantifying the uncertainties in our probability estimates and flagging pixels which have probability confidence estimates beyond our acceptable threshold as adversarial, all our comparisons of the baseline non-bayes model are against our bayesian \textit{Detection of Uncertainty in Exceedance of Threshold} (DUET) model.

\subsection{Settings}

We utilized different datasets and patch attack algorithms for training and evaluating our proposed adversarial patch localizer. We adopted the \textit{Flower102}~\cite{nilsback2008automated}, \textit{GTSRB}~\cite{stallkamp2012man} and the \textit{Food101}~\cite{bossard14} datasets. The \textit{Flower102} dataset consists of 8189 images of flowers belonging to 102 classes, while the \textit{GTSRB} dataset consists of 51839 images of german road signs belonging to 43 classes. The \textit{Food101} dataset consists of 101,000 image belonging to 101 classes. For our study, we utilized a subset of each dataset (4896 for training, 1632 for validation and 1632 for testing). We adopted several of the most renowned adversarial attack algorithms including the $l_{infinity}$-norm Projected Gradient Descent (PGD), Fast Gradient Sign Method (FGSM) and an adversarial patch, Google Patch.

During training, we trained the models on mini-batches of 8 adversarial images for 50 epochs on an RTX2080Ti GPU with an Adaptive Moment Estimation (ADAM) optimizer \cite{kingma2014adam} and learning rate of 0.1. We utilized mIoU (mean Intersection of Union) as our main evaluation metric as it is the most commonly used metric in the field of segmentation, and it works well in tackling class imbalance as it takes the average of the IoU of both classes, giving both classes equal weighting. Furthermore, mIoU is suited to handle class imbalance as experienced in our experiments. Although F1-score is recorded for our experiments, it is not reported as it is correlated with the mIoU score.

The models proposed were trained alongside the non-bayesian adversarial patch localizer on 4896 $l_{infinity}$-norm PGD attacked images from the Flower102 dataset. We have chosen to train on a $l_{infinity}$-norm PGD attacked image as PGD is considered one of the most renowned attack algorithms with proven attack successes. 

\subsection{Results and Discussion}
\begin{figure*}[t]
\centering
\subfigtopskip=2pt
\subfigbottomskip=2pt
\subfigure[Predicted adversarial patches by selected models on Flower102-PGD.]{
\begin{minipage}[t]{0.5\linewidth}
\centering
\includegraphics[width=0.9\linewidth]{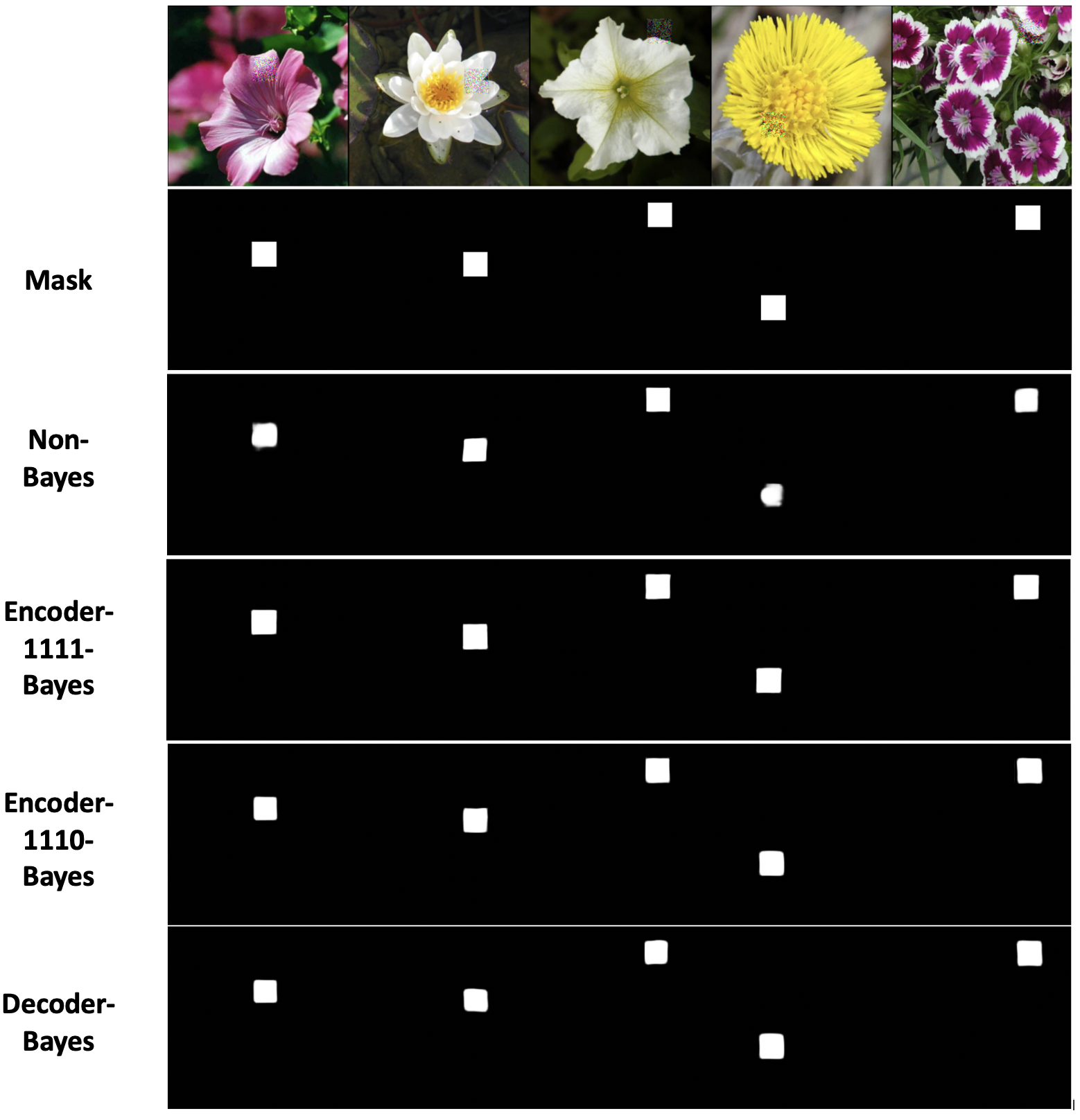}
\label{flower-pgd}
\vspace{-10mm}
\end{minipage}%
}%
\subfigure[Predicted adversarial patches by selected models on Flower102-Google.]{
\begin{minipage}[t]{0.5\linewidth}
\centering
\includegraphics[width=0.9\linewidth]{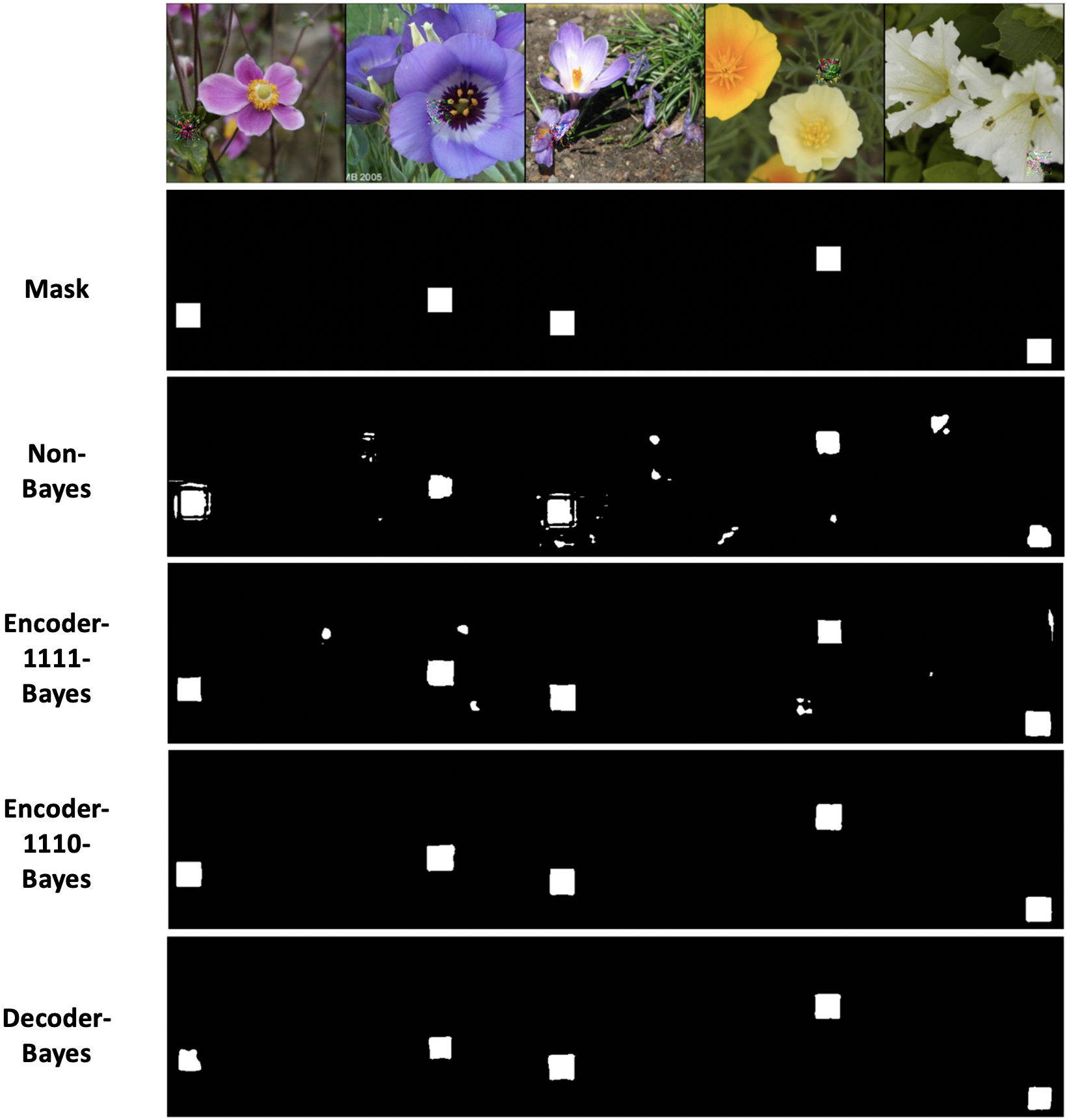}
\label{flower-google}
\vspace{-10mm}
\end{minipage}%
}%
\caption{Localization of PGD and FGSM attack patches, respectively.}
\vspace*{-0.5cm} 
\end{figure*}

\textbf{Test Set Performance. }The trained models were tested against a test set (PGD-attacked Flower102 dataset) and the DUET model (mIoU score = 0.921) outperformed the baseline model (mIoU score = 0.868) (shown in Table \ref{pgdtable}). From the output predicted adversarial patches, the DUET model showed highly accurate patch localization and defined patch boundaries. In contrast, the baseline model frequently showed signs of underestimation of the adversarial patch, with predicted patches smaller than the true patch size, with jagged corners or undercut edges (shown in Figure \ref{flower-pgd}).

\begin{figure*}[t]
\centering
\subfigtopskip=2pt
\subfigbottomskip=2pt
\subfigure[Predicted adversarial patches by selected models on GTSRB-FGSM.]{
\begin{minipage}[t]{0.5\linewidth}
\centering
\includegraphics[width=0.9\linewidth]{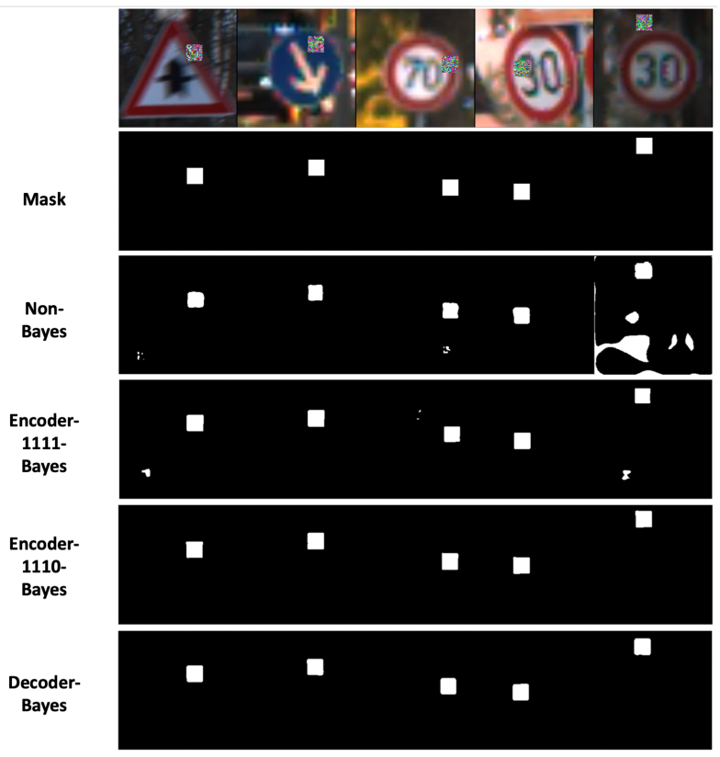}
\label{gtsrb-fgsm}
\vspace{-10mm}
\end{minipage}
}%
\subfigure[Predicted adversarial patches by selected models on Food101.]{
\begin{minipage}[t]{0.5\linewidth}
\centering
\includegraphics[width=0.9\linewidth]{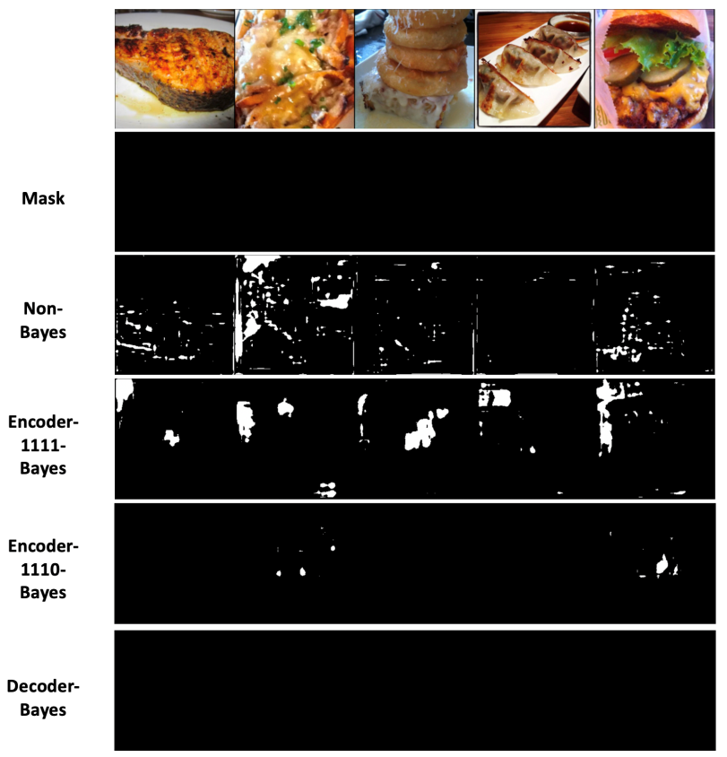}
\label{food-clean}
\vspace{-10mm}
\end{minipage}
}%
\caption{Localization of FGSM attack patches and attack patches in absence of attack, respectively.}
\vspace*{-0.5cm} 
\end{figure*}

\textbf{Performance with different priors. } We found that the model which used the Cauchy prior (mIoU = 0.924) outperformed the model which used the GMM prior (mIoU = 0.850) and exhibited comparable performance to the standard isometric gaussian (mIoU = 0.921).

To ensure the reliability of the DUET model in detecting adversarial patches, we test the generalizability of our model to alternative datasets with other forms of attack algorithms. We compared the Cauchy-prior model with the isometric gaussian-prior model due to their comparable performance in localizing adversarial patches on the standard test dataset. We tested both models on the \textit{Flower102} dataset attacked by \textit{Google Patch} and on the \textit{GTSRB} dataset attacked by \textit{FGSM} algorithm and found that the isometric gaussian prior model produced a better average performance (mIoU = 0.654, standard deviation = 0.1; mIoU = 0.937, standard deviation = 0.0169, respectively) when compared to the Cauchy-prior model (mIoU = 0.525, standard deviation = 0.35; mIoU = 0.497, standard deviation = 0.349, respectively) (shown in Table \ref{fulltable}). The Cauchy-prior model is unreliable as it produced refined adversarial patch predictions for some images but completely overestimated the size of the adversarial patch by classifying large portions of the image as part of an adversarial patch. This is in contrast to the findings by \cite{gillsjo2021depth} which found that Cauchy-prior improved their model performance. Nevertheless, our experiment outcomes are in line with \cite{fortuin2021bayesian} which found that gaussian priors were suitable, albeit a multivariate gaussian, for 2D convolution kernels.

\textbf{Selection of Bayesian layers. }Bayesian neural networks have been widely used in existing literature \cite{deng2021libre,li2021detecting} within the subfield of adversarial detection. Works such as \cite{deng2021libre} have only adopted the last few fully connected layers as Bayesian neural nets for the purpose of classifying adversarial images. In \cite{li2021detecting}, they converted every layer in their network into Bayesian for the purpose of adversarial detection via divergence of adversarial example's hidden layers' distribution from natural examples. However, there is no existing study, to the best of our knowledge, which studied the implications and performance of applying bayesian neural networks at different layers of a segmentation model. In a traditional segmentation model, there are typically the encoder and the decoder section of the model. Using the \textit{Deeplab V3+} segmentation architecture with a \textit{ResNet-101} backbone~\cite{he2016deep}, we explore the implications and performance of bayesian neural networks in different sections of the model.

In our setup, we applied our bayesian neural nets according to the blocks of a \textit{ResNet-101}; block-1, block-2, block-3, block-4. We considered several encoder-only block combinations of bayesian neural layers, a fully bayesian neural network, and a decoder-only bayes network. We studied how these models compare to one another, and to the baseline non-bayes model. In our first inference test, using the test dataset (Flower102-PGD) which has similar distribution to the training set, we note that most DUET models performed better than the non-bayes baseline model (mIoU = 0.868), with the exception of the fully bayes DUET model (mIoU = 0.208) which failed to converge, and encoder-0001-bayes (encoder-last block-bayes) DUET model (mIoU = 0.784). We postulate that a completely bayes DUET model has too many stochastic layers and may require significantly more epochs of training to converge. As for the encoder-0001-bayes DUET model, we note that implementing bayesian layers in the final block of the encoder resulted in degradation of model performance on the test set. On the other hand, the encoder-0111-bayes, encoder-1000-bayes, encoder-1100-bayes, encoder-1110-bayes, and the encoder-1111-bayes DUET model achieved the best test set performances (mIoU = 0.926, 0.920, 0.921, 0.934, 0.921, respectively). Sample images and predictions are shown in Figure \ref{flower-pgd}. Results are shown in Table \ref{pgdtable}. Nevertheless, these performance scores are based on a test set with similar distribution to the training set. 


\begin{table}[tp]
\centering
\caption{mIoU scores on Flower102-PGD for selected models.}\label{pgdtable}
\vspace{0.2mm}
\begin{tabular}{cc}\hline
 & Test mIoU (PGD) \\ \hline
Non-Bayes   & 0.868 \\ \hline
All-Bayes-Gaussian   & 0.209 \\ 
Encoder-Bayes-Cauchy   & 0.924 \\ 
Encoder-Bayes-Gaussian   & 0.922 \\ 
Encoder-0001-Bayes   & 0.784 \\ 
Encoder-0010-Bayes   & 0.899 \\ 
Encoder-0011-Bayes   & 0.887 \\ 
Encoder-0100-Bayes   & 0.884 \\ 
Encoder-0110-Bayes   & 0.908 \\ 
Encoder-0111-Bayes   & 0.926 \\ 
Encoder-1000-Bayes   & 0.921 \\ 
Encoder-1100-Bayes   & 0.922 \\ 
Encoder-1110-Bayes   & 0.935 \\ \hline
Decoder-Bayes   & 0.870 \\ \hline
\end{tabular}
\vspace*{-0.5cm} 
\end{table}

\textbf{Unseen Attack Set Performance. }In order to observe the models' generalizability to unseen attacks and dataset, we test them at inference time against two different datasets: (i) Flower102 images attacked by google patch (Flower102-Google) and (ii) GTSRB images attacked by Fast Gradient Sign Method (GTSRB-FGSM). From the inference results, we observed that while most of the DUET models perform significantly well on GTSRB-FGSM, many of them showed poor predictive performance on the Flower102-Google. Most notably, the encoder-1111-bayes, encoder-0111-bayes, encoder-1000 DUET models which achieved stellar test performance on the original test set generalized poorly (obtaining mIoU = 0.654, 0.555, 0.678, respectively) on an attacked image with the Google Patch attack. These models had unreliable performances, predicting and localizing adversarial patches very well for some images and completely faltering at others, as shown by their test standard deviations (0.105, 0.220, 0.109, respectively). These models showed signs of high uncertainty and consequently overestimated the adversarial patch area drastically. On the other hand, previously top-performing encoder-1100-bayes and encoder-1110-bayes DUET model displayed good performance and high confidence on Google Patch attack (mIoU=0.857, s.d=0.0461; mIoU=0.906, s.d.=0.0234, respectively) and on the FGSM attack (mIoU=0.919, s.d.=0.0103; mIoU=0.958, s.d.=0.00628, respectively). It is also important to note that previously relatively modest performing decoder-only-bayes and encoder-0011-bayes DUET models held up and performed consistently well on the Flower102-Google (mIoU=0.868, s.d.=0.0233; mIoU=0.808, s.d=0.0682, respectively) and on the GTSRB-FGSM (mIoU=0.964, s.d.=0.00584; mIoU=0.948, s.d.=0.00671, respectively), showing great generalizability to alternative attacks and dataset. Nevertheless, the above bayesian DUET models performed at minimal, better and much more reliably than the non-bayes adversarial patch localizer baseline model on the Flower102-Google (mIoU=0.518, s.d.=0.342) and on the GTSRB-FGSM (mIoU=0.773, s.d.=0.233). The unreliability and mis-predictions of the non-bayes model include underestimating the patch area in certain images, or severely overestimating the patch area, confusing the images' contours or defining object boundaries with adversarial patches. Sample images and predictions are shown in Fig. \ref{flower-google}, and Fig. \ref{gtsrb-fgsm}. Results are shown in Table \ref{fulltable}.

\begin{table*}[tp]\small
\centering
\caption{mIoU scores on Flower-Google, GTSRB-FGSM, Food101-Clean for selected models.}\label{fulltable}
\vspace{0.2mm}
\scalebox{1}{
\begin{tabular}{ccccccc}
\hline
 & Google Patch (mean) & Google Patch (s.d.) & FGSM (mean) & FGSM (s.d.) & Clean (mean) & Clean (s.d.) \\ \hline
  \multicolumn{7}{c}{Non-bayes} \\
\hline
\textbf{Non-Bayes}   & 0.519 & 0.342 & 0.773 & 0.234 & 0.854 & 0.336 \\ \hline
  \multicolumn{7}{c}{Encoder-bayes} \\
\hline
All-Bayes-Gaussian   & 0.087 & 0.051 & 0.883 & 0.017 &  &  \\
Encoder-Bayes-Cauchy   & 0.526 & 0.355 & 0.0497 & 0.349 & 0.926 & 0.070 \\
Encoder-Bayes-Gaussian   & 0.654 & 0.106 & 0.938 & 0.017 & 0.917 & 0.021 \\
Encoder-0001-Bayes   & 0.516 & 0.124 & 0.930 & 0.012 &  &  \\
Encoder-0010-Bayes   & 0.854 & 0.049 & 0.929 & 0.015 & 0.937 & 0.016 \\
Encoder-0011-Bayes   & 0.808 & 0.068 & 0.948 & 0.007 & 0.998 & 0.001 \\
Encoder-0100-Bayes   & 0.588 & 0.067 & 0.943 & 0.008 &  &  \\
Encoder-0110-Bayes   & 0.768 & 0.169 & 0.784 & 0.249 & 0.993 & 0.005 \\
Encoder-0111-Bayes   & 0.555 & 0.221 & 0.919 & 0.015 &  &  \\
Encoder-1000-Bayes   & 0.678 & 0.110 & 0.952 & 0.005 &  &  \\
Encoder-1100-Bayes   & 0.858 & 0.046 & 0.920 & 0.010 & 0.969 & 0.010 \\
\textbf{Encoder-1110-Bayes}   & 0.906 & 0.023 & 0.959 & 0.006 & 0.990 & 0.004 \\ \hline
  \multicolumn{7}{c}{Decoder-bayes} \\
\hline
\textbf{Decoder-Bayes}   & 0.869 & 0.023 & 0.965 & 0.006 & 1.000 & 0.000 \\
\hline

\end{tabular}
}
\vspace*{-0.5cm} 
\end{table*}

\textbf{Clean Image Set Performance. }Finally, to ascertain a detector model does not over-fit to the adversarial training set, we observe its predictive performance to clean (non-adversarial) images (Food101 Dataset). In attempting to select the overall best performing model(s), we only included models which performed relatively well on the original test images and generalized relatively well on the alternative attacks and dataset. Nevertheless, we include the non-bayes baseline adversarial patch detector model for comparison. Note that the mIoU metric we use in this inference test only considers the successful segmentation of the background (not considering the segmentation of the adversarial patch) as mIoU takes into account the mean IoU of both classes and several models perfectly segments an image, resulting in no pixels being classified as adversarial, rendering the calculation of mean IoU erroneous). From the test result, we observe that the non-bayes adversarial patch localizer performed significantly poorer, on average (mIoU=0.854, s.d.=0.336) than the bayes DUET models in segmenting clean images. This poor result is a consequence of an unreliable segmentation model. In several instances, the non-bayes model achieves near-perfect segmentation of clean images. However, in some instances, the non-bayes model confuses large portions of the image with the adversarial patch, and in some other instances, the model classifies parts of the image as adversarial, with no visibly discernible shape. Comparing between DUET models, result show that the encoder-1111-bayes (mIoU=0.916, s.d.=0.0213) and encoder-0010-bayes (mIoU=0.936, s.d.=0.0159) DUET model segmented clean images relatively poorer in contrast to the other tested DUET models. These models acted conservatively and misclassified significant portions of some clean images as adversarial. These misclassified adversarial patches do not resemble any discernable shapes. The encoder-0011-bayes and encoder-1110-bayes models show remarkable segmentation performance and reliability (mIoU=0.998, s.d.=0.00132; mIoU=0.989, s.d.=0.00366) on clean images, with occasional globules of mispredicted adversarial patches. The decoder-bayes DUET model achieves the best performance among the tested models (mIoU=0.999, s.d.=1.99E-06) with an almost perfect segmentation on the clean image. Sample images are shown in Figure \ref{food-clean}.

\textbf{Analyses on Bayesian layers. }From test results, we observe that the non-bayes adversarial patch localizer does not achieve as good performance and generalizability as its Bayesian DUET counterparts. One possible rationale is that the point estimate weights in traditional neural networks may over-fit and learn specific characteristics and representations from the training set. For the DUET models, bayesian layers on single encoder blocks do not provide the model with substantial generalization ability. This can be shown by the weaker performance of these models on the Flower102-Google and the GTSRB-FGSM. This characteristic is more obvious with bayesian layer being on the last block of the encoder producing significantly poorer performance and generalizability. Bayesian neural networks on two or more blocks seem to improve the model's generalizability. This signals a possibility that more bayesian blocks could be necessary in deep learning models to improve a model's generalizability to the unseen dataset. Another interesting observation is that out of the 4 ResNet-101 blocks, the addition of bayesian layers to the first two blocks seems to facilitate better top-end performance, achieving much higher predictive scores on detecting adversarial patches. In contrast, the presence of bayesian layers in the last few blocks in the encoder seems to facilitate better stability in clean image prediction performance. However, it seems that placing bayes layers on all blocks in the model's encoder, or even on the entire segmentation model leads to significant degradation in model convergence. This could be due to excessive stochasticity which makes training difficult. Fully bayesian layers on the segmentation model's decoder have shown very rounded performance, achieving good performance on the test set, alternative attacks (albeit not the best), and yet have shown remarkable accuracy and confidence in predicting clean images. This is in contrast to an encoder-1110-bayes model which achieved top-end performance on test and alternative attack set, but was overestimating the adversarial patch attacks on the clean image. Ultimately, we conclude that the two best DUET models from our study are the encoder-1110-bayes model and the decoder-only-bayes model. However, a decoder-only-bayes model would incur less inference latency than an encoder-bayes model. The rationale for a greater inference latency in an encoder-bayes system is that multiple forward passes through the entire model is required to obtain an output probability distribution, while a decoder-bayes model could use one forward pass through the deterministic layers in the model and only execute the multiple forward passes in the decoder section of the model. If achieving top-end predictive performance in capturing and localizing adversarial patches is essential, while inference latency and over-estimating the size and presence of adversarial is not detrimental, encoder-1110-bayes model would be desirable. On the other hand, if the application demands a rounded performance with great non-adversarial image performance, the decoder-only-bayes model would be a better selection. It should also be noted that we did not achieve building a DUET model which has both top-end adversarial patch localization performance and near-perfect clean image predictive performance, concurrently. The addition of more bayesian layers in our study did not improve performance but still degraded it. There could be a trade-off between achieving top-end adversarial performance in adversarial patch localization and clean image performance and demands further work. In addition to that, we have observed that a decoder-only-bayes model significantly outperforms some other encoder-bayes models we have tested. Further work into understanding the configurations and implementations of bayes layers in the decoder of segmentation models should be explored.

\section{Conclusion} 
\label{section:conclusion}
In our work, we have  developed a novel uncertainty-based adversarial patch localizer which contributes to the field of adversarial patch localization. We have also proposed the \textit{Detection of Uncertainty in the Exceedance of Threshold} (DUET) algorithm which ascertains confidence in our prediction estimates of adversarial pixels. This adversarial patch localizer under the DUET framework opens doors for adversarial patch detection and localization in safety-sensitive applications. We also provided in-depth analyses on the selection of model priors, on the placement of bayesian layers in segmentation models and produced two well-performing models for different use cases.

\section*{Acknowledgement}

This research is supported in part by Nanyang Technological University (NTU) Interdisciplinary Graduate Program (IGP); the NTU-Wallenberg AI, Autonomous Systems and Software Program (WASP) Joint Project; NTU Startup Grant; the Singapore Ministry of Education Academic Research Fund under Grant Tier 1 RG97/20, Grant Tier 1 RG24/20 and Grant Tier 2 MOE2019-T2-1-176.

\bibliographystyle{IEEEtran}
\bibliography{sample-base}

\end{document}